%% file: sample_new.tex
\documentclass[letterpaper, 10 pt, conference]{ieeeconf}  

\IEEEoverridecommandlockouts                              
\usepackage{epsfig} 
\usepackage[applemac]{inputenc}
\usepackage{epsf}
\usepackage{amsmath} 
\usepackage{subfigure}

\title{\LARGE \bf
Gaussian Mixture Models for Affordance Learning\\
using Bayesian Networks
}

\author{Pedro Os\'orio \and Alexandre Bernardino \and Ruben Martinez-Cantin \and Jos\'e Santos-Victor
\thanks{This work was partially supported by the Portuguese Government - Funda\c c\~ao para a Ci\^encia e Tecnologia (ISR/IST pluriannual funding) through the POS Conhecimento Program that includes FEDER funds, and through  the EC Funded projects HANDLE and FIRST-MM.
}
\thanks{P. Os\'orio, student of Biomedical Engineering and collaborator of the Institute for Systems and Robotics, IST, Lisboa, Portugal
        {\tt\small pedro.osorio@ist.utl.pt}}%
\thanks{A. Bernardino, R. Martinez-Cantin and J. Santos-Victor are with the Institute for Systems and Robotics, IST, Lisboa, Portugal
        {\tt\small \{alex,rmcantin,jasv\}@isr.ist.utl.pt}}%
}

\begin{document}

\maketitle
\thispagestyle{empty}
\pagestyle{empty}

\begin{abstract}

Affordances are fundamental descriptors of relationships between actions, objects and effects. They provide the means whereby a robot can predict effects, recognize actions, select objects and plan its behavior according to desired goals.  
This paper approaches the problem of an embodied agent exploring the world and learning these affordances autonomously from its sensory experiences. Models exist for learning the structure and the parameters of a Bayesian Network encoding this knowledge. Although Bayesian Networks are capable of dealing with uncertainty and redundancy, previous work considered complete observability of the discrete sensory data, which may lead to hard errors in the presence of noise. In this paper we consider a probabilistic representation of the sensors by Gaussian Mixture Models (GMMs) and explicitly taking into account the probability distribution contained in each discrete affordance concept, which can lead to a more correct learning. 

\end{abstract}

 \input{intro}

 \input{body}

 \input{results}

 \input{concfuture}

\section{ACKNOWLEDGMENTS}

This work has been partly supported by PTDC/EEA-ACR/70174/2006, grant SFRH/BPD/48857/2008 from FCT and EU projects HANDLE and FIRST-MM.

\input{appendix}

\bibliographystyle{plain}
\bibliography{pratz,../bibs/BN,../bibs/macl,../bibs/affordance,../bibs/manifolds}

\end{document}

%% file: intro.tex
\begin{section}{Introduction}
Affordances define the relationships between actions, objects and effects.
As defined by James J. Gibson more than 30 years ago \cite{gibson79ecological} affordances are \emph{agent-dependent object usages}, i.e. the action possibilities given by an object to an agent with specific action capabilities. For example, a chair is only \emph{sitable} by an individual of a certain height; a tree is only \emph{climbable} by animals with specific capabilities. In fact, what matters in obtaining a certain desired effect are the properties of the objects more than the objects themselves.

The knowledge of object affordances is exploited in most of our decisions: to choose the most appropriate way of acting upon an object for a certain purpose; to search and select objects that best suit the execution of a task; to predict the effects of actions on objects; to recognize ambiguous objects or actions; etc. Affordances are at the core of high-level cognitive skills such as planning, recognition, prediction and imitation. 

Thus, learning object affordances is an essential step for humans as they develop the required skills to interact with the environment and ultimately with each other. It is logical that affordances would be modeled in learning humanoid robots, emulating the human development process.

A few works have addressed the problem of learning affordances by autonomous exploration of the world. In \cite{metta03poking}, a robot learned the direction of  motion of different objects when poked and used this information at a later stage to recognize actions performed by others. In \cite{macl08affTRO} a set of predefined actions (tap, touch, grasp) are applied to objects of different shapes, colors and sizes, and the observed effects (object/hand velocities, contact) are used to learn a network of cause-effect relationships. However, these works do not address the noisy characteristics of the perceived sensory data, which may lead to difficulties in the learning stages. This work builds on~\cite{macl08affTRO} but extends it in order to model the learning process under noisy observations. 

The paper is organized as follows. In the next section we present the affordance learning methodology in \cite{macl08affTRO} where we base our work. Then, we describe our main contribution: to model the noisy nature of observations and the derivation of an EM algorithm for learning the parameters of the Bayesian Network. We present results from simulations where we can observe the improvements provided by the proposed methodology. Finally, we draw the conclusions and point directions for future work.
\end{section}
\begin{section}{Learning Object Affordances}

In this section we briefly describe the affordance learning process as described in \cite{macl08affTRO}. It consists in a developmental approach composed by three stages: (i) learning motor primitives, (ii) learning object representations, (iii) learning effects representation and finally (iv) learning the object's affordances.
\begin{subsection}{Developing Basic Skills}
The first skills learned by the robot are the motor ones. It starts with a number of predefined actions such as grasping, tapping or touching and fine-tunes these, using its sensory abilities (i.e. relating the observed motion with the desired motion).
After this, the robot is presented with different objects and forms perceptual categories. That is, for a number of properties of an object (color, shape, size) it uses unsupervised learning to form meaningful clusters (e.g. it may find 3 clusters in the size space -- small, medium, big).
In a subsequent stage the robot interacts with objects and, just as in the case of visual perception, it forms clusters for a number of properties (object velocity after contact, contact duration, etc.). Finally, when the robot has learned to execute actions (discrete) and categorize the objects and effects features in discrete categories, it is ready to start creating cause-effect links between actions on objects and the particular outcomes. It is on this stage that we concentrate our attention.
\end{subsection}
\begin{subsection}{Affordance Modeling and Learning}
\label{sec:afflearn}
Based on the work of \cite{macl08affTRO}, we use a graphical model known as Bayesian Network (BN) to model affordances. This model is used to encode probabilistic dependencies between actions, object features and effects of the actions. In this graphical model, nodes represent random variables and the arc structure encodes conditional independence assumptions between the variables. A fundamental result of BN is that the joint distribution of the variables decomposes in the following way: 
\begin{equation} 
  p(X_1,...,X_n) = \prod_{i=1}^n p(X_i | X_{Pa(X_i)}, \theta_i)\label{joint}
\end{equation}
where $X_i$ is the random variable associated with node $i$, $\theta$ is the set of parameters of the probability distribution and $Pa(X_i)$ are the parents of node $i$.

Figure~\ref{graph1} is an example of a BN with the characteristics of the model used in this work. All random variables are observed and discrete-valued. $a_i$ are the actions, $o_i$ are the object features and $e_i$ are the effects of the action.

\begin{figure}
\centering
\includegraphics[width=0.5\columnwidth]{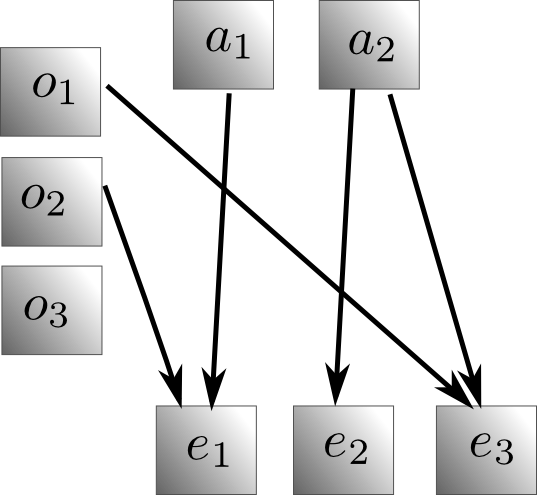}
\caption{BN with discrete,observed nodes\label{graph1}}
\end{figure}

Learning the affordances requires that, given a database of cases (observations of actions, objects and effects), we find the structure of the BN and the parameters $\theta _i$ that describe the relationships between each node and its parents. The structure can be efficiently approximated using Markov Chain Monte Carlo (MCMC) as well as the K2 algorithm.

Since this model only has discrete, observed nodes, we can model the conditional probability distributions of each node given its parents as a multinomial distribution and we can use the Dirichlet distribution as the prior distribution. Then, we can use conjugate Bayesian analysis to do the inference  and update the parameters (see~\cite{Heckerman1996}).
\end{subsection}
\begin{subsection}{Inference and Interaction Games}
After the affordances are learned it is possible to use this knowledge during execution to infer incomplete data using a Bayesian inference strategy. For example, this can be used to interact with others. If a human performs an action on a given object, the robot can observe the effect. Even without a mechanism to identify actions carried out by others, the robot can use the object features, the effect and the learned affordances to infer the action. In fact, we can use the model to predict or infer any of the variables (actions, objects or effects) given the other two.
\end{subsection}
\end{section}

%% file: body.tex
\section{Probabilistic clustering for Affordances}
The affordances model is based on the assumption that the variables are separable, because objects features and effects must have some semantic meaning. In some sense, it is the first step towards relational learning, where abstract classes define the properties for interaction. Therefore, the first step is to find the semantic classes by unsupervised learning. In the work of \cite{macl08affTRO}, this clustering is based on \emph{X-means} \cite{pelleg-xmeans}, which represents the data distribution as a Gaussian Mixture Model. However, the work of \cite{macl08affTRO} assigns the \emph{maximum a posteriori} (MAP) cluster to each data point. Intuitively, this is a strong assumption considering many of the variables that are involved in the affordances model, like object features such us \emph{color} or \emph{size}. 

For example, let us consider that our unsupervised learner defines a ball with a radius greater than 5 cm as \emph{large}, and \emph{small} otherwise. When the sensor detects an object with a radius of 4.9 cm, it automatically assign the semantic meaning of \emph{small} object. Intuitively, we can see that the object in fact falls in between. If we consider the sensor noise, there is a high probability of the object being \emph{small}, but also there is a non-negligible probability of being \emph{large}. In our approach, we want to give a probabilistic assignment to the different classes, which is more robust and can be used to do Bayesian inference in the semantic level.

The Gaussian Mixture Model obtained from \emph{X-means} and similar algorithms provides a density estimation of the probability distribution of the data. The advantage of using mixture models is that we can still assign a semantic meaning to every component, while recovering the whole distribution for Bayesian inference. In the \emph{size} example, we may find that the distribution is the combination of two Gaussians, such us $\mathcal{N}(4,1)$ for the \emph{small} objects and $\mathcal{N}(6.5,2)$ for the \emph{large} ones. Therefore, our previous object of radius 4.9 cm has a probability of being \emph{small} and a non-negligible probability of \emph{large}. It is important to note that the inference model that we define in this paper could also deal with probabilistic inputs from the actions. However, in the robotic setup, control noise is negligible compared to sensor noise. Therefore, we consider deterministic actions to simplify the explanation.

\begin{figure}
\centering
\includegraphics[width=0.6\columnwidth]{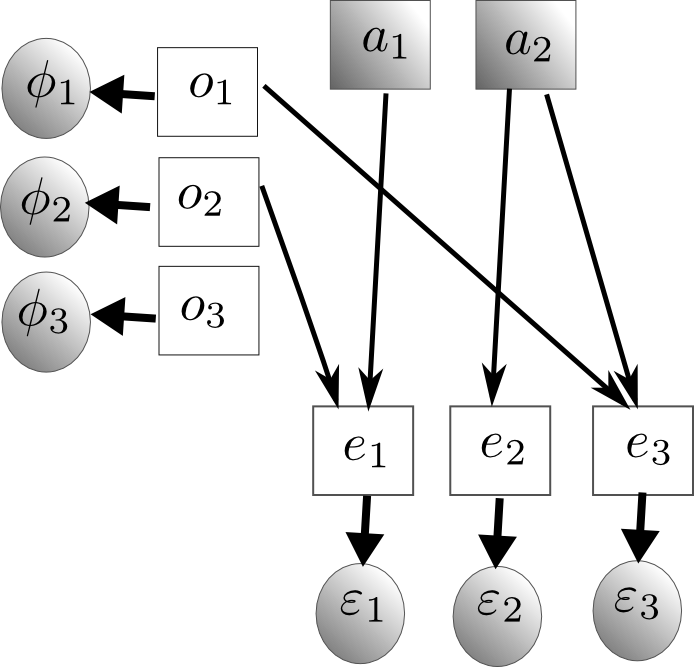}
\caption{Bayesian Network of the new clustering approach. The affordances are still represented as interactions between discrete variables (squares). The GMM is represented by the new continuous variables (circles). Shaded shapes represents observed variables during training $\mathcal{X} = (Actions, Objects, Effects)$. However, during execution, we can estimate any variable of $\mathcal{X}$, given the other two observations. The thick arrows represent the new dependence of the sensors and their associated features}
\label{graph2}
\end{figure}

Having defined a probabilistic assignment of the clusters to the sensing variables, we can extend the BN to consider the new dependencies (Fig.~\ref{graph2}). The nodes that represent the discrete classes of object features and effects are no longer observed directly. Instead, the observed nodes are their children, the sensor nodes which represents continuous random variables.

\subsection{Learning the Structure}
By definition, we know that there is a dependency between any continuous node -sensor- and the corresponding discrete variable -feature-. Besides, the sensor is independent of the rest of the network given the value of the feature. Therefore, we can assume that this part of the structure is known and fixed, simplifying the structure learning problem. In fact, the structure to be learned is only the one that connects discrete nodes, like in \cite{macl08affTRO}. However, the problem is different because, in this case, some of the nodes are hidden. We can use a Bayesian method such us Structural-EM \cite{Friedman98}, which allows to estimate the structure of the network by integration over all possible hidden variables.  However, this Bayesian approach is computationally very expensive and typically requires large datasets. Instead, as in \cite{macl08affTRO} we assign the MAP value to any hidden variable. The MAP value is the one that corresponds to
\begin{align}
\begin{split}
 e_i^* &= \arg \max_{e_i}  p(\varepsilon_i|e_i) \cdot p(e_i)\\
 o_i^* &= \arg \max_{o_i}  p(\phi_i|o_i) \cdot p(o_i)
\end{split}
\end{align}
where the likelihood $p(\varepsilon_1|e_1)$ or $p(\phi_i|o_i)$, and the prior $p(e_i)$ or $p(o_i)$, come from the estimation of the parameters of the GMM.

Once we have assigned the MAP value to the discrete variables -features-, we can treat them as observed variables. Therefore, we can apply the structure learning algorithms used the work of \cite{macl08affTRO}. As commented in section~\ref{sec:afflearn}, we can use MCMC \cite{Madigan95} and K2 \cite{Cooper92} to learn the structure as if all nodes were observed. An important issue when doing structure learning appears when dealing with causality. In order to find causal relationship, we need to use interventional data. However, in robot interaction, causality is implied in the variables, that is, \emph{acting on an object produces and effect}.

\subsection{Learning the Parameters}

Once we have learned the structure, we can proceed to learn the parameters of the BN as commented in section~\ref{sec:afflearn}. However, our new model has to deal with incomplete data from the sensors, that is, in terms of network, some variables are hidden. Therefore, we need to integrate over all the hidden variables. For that purpose, we can use the EM algorithm \cite{dempster77em}.

Before describing in the detail how this is done for the previously described model, the notation used is the following: 
\begin{itemize}
 \item Uppercase letters are used to refer to variables, and lowercase to their values. However, we will often abuse of the notation and use the lowercase both for the variables and their respective values.
 \item  $Y = (Y^{(1)},\dots,Y^{(k)})$ is the set of observed variables, where the superindex represent the number of experiment $k$.
 \item  $Z = (Z^{(1)},\dots,Z^{(k)})$ is the set of hidden variables, where the superindex represent the number of experiment $k$. For each experiment, $Z^{(k)}_{i}$ being the $i^{th}$ hidden node, where $i=1,\dots,N$.
 \item  $X = (X^{(1)},\dots,X^{(k)})$ is the set of variables for an experiment, being $X = Y \cup Z$.
 \item $Pa(Z_i)$ is the set of the parents of $Z_{i}$ and $\overline{Pa}(Z_i)$ is the set of hidden variables that are not parents of $Z_{i}$. Note that the structure is independent of the experiment, so $Pa(Z_i)$ and $\overline{Pa}(Z_i)$ are independent of $k$.
 \end{itemize}

Since the log-likelihood of the parameters given the complete data, $\log[ p(X|\theta)]$ cannot be computed (due to partial observability), the E-step computes its expected value with respect to the distribution $p(Z | Y, \theta^{old})$, where $\theta^{old}$ is the vector of parameters achieved on the last iteration of EM. The set of parameters $\theta$ that maximize this expected value are then computed in the M-step. In order to simplify calculations, we also assume that the action is only observed through a sensor. Therefore, there is also an auxiliary hidden node for the action, with the corresponding observed child node. This is equivalent to an observed action node but allows the separation of the observed nodes from the hidden ones on the joint probability of the network \eqref{joint}. 

\subsubsection{E step} -- Compute the expected value of the complete log-likelihood with respect to the distribution of the complete data (on $K$ experiments):

\begin{equation}
Q(\theta,\theta^{old}) = \alpha + \sum_{k=1}^K \sum_{z_k} \left[ w_k \sum_{i=1}^N \log \theta^{(k)}_{i} \right] \label{loglike}
\end{equation}
where
\[
 \alpha = \sum_{k=1}^K \sum_{z_k} \left[ w_{k} \cdot \log p(Y^{(k)}|Z^{(k)} \right] 
\]
being
\[
 w_k = p(Z^{(k)}|Y^{(k)},\theta^{old})
\]
and
\[
 \theta_i = p(Z^{(k)}_{i}|Pa(Z_{i}),\theta)
\]

So the E-step consists of computing $w_k$. It should be noted that $w_k = w(Z^{(k)})$ and $\sum_{z_k} w(Z^{(k)}) = 1$. The derivation of equation~\eqref{loglike} can be found in the appendix.

\begin{figure*}
\centering
\subfigure[]{\includegraphics[width=0.7\columnwidth]{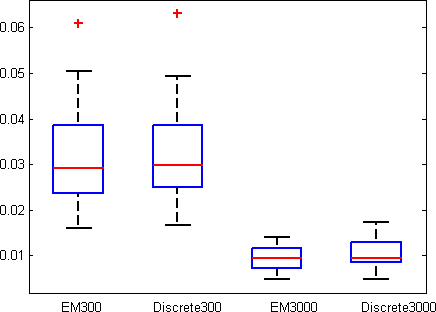}\label{res1}}
\hspace{8pt}%
\subfigure[]{\includegraphics[width=0.7\columnwidth]{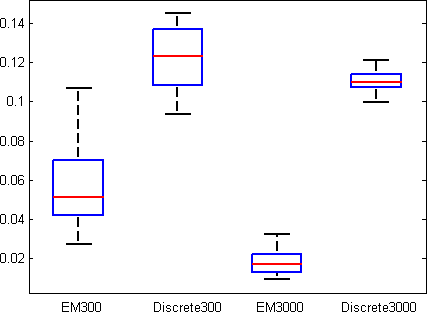}\label{res2}}
\hspace{8pt}%
\subfigure[]{\includegraphics[width=0.5\columnwidth]{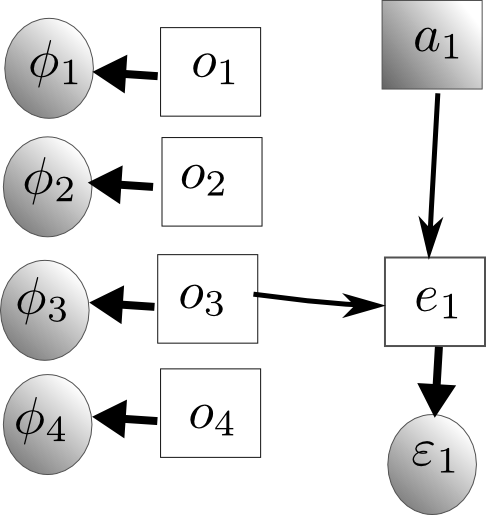}\label{graph4}}
\caption{RMS error of the difference between learned parameters and the real ones given a database with 300 or 3000 cases. a) Sensor noise distributed according to $\mathcal{N}(0,1)$, b) Sensor noise distributed according to $\mathcal{N}(0,9)$ c) The simple Bayesian network.}
\end{figure*}

\subsubsection{M step} -- perform the maximization 
\[\theta^* = \arg \max_\theta \ Q(\theta,\theta^{old})\]
subject to the constraint
\[
\sum_{z_{i}=1}^{r_i} \theta_i = 1
\]
where $r_i$ is the number of different values the $i^{th}$ node can take. Basically this restriction corresponds to the partition function, that is, for a fixed configuration of its parents, the sum of the probabilities for all possible realizations has to be $1$. The optimal value for $\theta_i$ is:
\begin{equation} 
\theta_i =\frac{\sum_{k=1}^{K} p(z^{(k)}_i,Pa(z_i)|Y^{(k)},\theta^{old})}{\sum_{z=1}^{r_i} \sum_{k=1}^{K} p(z^{(k)}_i,Pa(z_i)|Y^{(k)},\theta^{old})} \label{theta}
\end{equation}
If we consider the probability $p(z^{(k)}_i,Pa(z_i)|Y^{(k)},\theta^{old})$ as the frequency of appearance of the realization, the solution is analogous to the solution of the complete-data problem. On the fully observable case (not considering the priors) we assigned:
\begin{equation} 
\theta_i = \frac{\sharp(z_i,Pa(z_i))}{\sharp(Pa(z_i))} 
\end{equation} 
where $\sharp(z_i,Pa(z_i))$ is the number of cases of the database that have $Z_i = z_i$ as well as $Pa(Z_i) = Pa(z_i)$ and $\sharp(Pa(z_i)) = \sum_{z_i=1}^{r_i} N(z_i,Pa(z_i))$.

With hidden-variables, instead of occurrence (1) or non-occurrence (0) of a specific configuration $z^{(k)}_i$ , $Pa(z_i)$ on each case of the database, we have the probability of this configuration given the observed variables and the parameters of the network computed on the last step. The \emph{counts} are replaced by the sum of these probabilities over all the cases of the database.

This algorithm \emph{as is}, has a time complexity which is linear on the number of cases in the database, and on the number of iterations. It is, however, exponential on the number of nodes.

%% file: results.tex
\begin{section}{Results}

\begin{subsection}{Procedure}

In order to evaluate the differences between the first approach and the one proposed here, two types of Bayesian networks were used. The first (Fig.~\ref{graph4}) is a sparse graph (only two arcs, going to the same variable) with seven discrete variables. Each has two possible values which leads to 10 degrees of freedom in the parameters. The second one (Fig.~\ref{graph3}) is the Bayesian network learned on ~\cite{macl08affTRO}. It has eight variables and ten arcs, the discrete variables have between two and four possible values, the parameters of this model have 125 degrees of freedom. 

Each of the discrete nodes in these Bayesian networks has a sensor node, similar to the ones described earlier. The purpose of this work is to learn the parameters of the discrete network, assuming that we know the distributions of the sensor models. Thus, the conditional distribution of each sensor given the value, $v$, of the associated discrete variable was arbitrarily defined to be a normal with mean $5 \cdot v$, and variance constant for all $v$\footnote{Here, $v$ is assumed to take all integer values between 0 and n-1, if the discrete variable has n possible values}.

In order to test the EM algorithm against the discrete representation used in \cite{macl08affTRO}, hereinafter called \emph{Discrete}, we generate a database of \emph{simulated experiments} based on a known sensors model and Bayesian network, which we will use as ground truth. In order to avoid errors from the clustering algorithm, we assume that the clusters are known. The \emph{Discrete} case takes the most probable cluster for each data point. The EM algorithm is applied to the data and let run until the root mean square (RMS) error of the difference between the parameters obtained in consecutive iterations is smaller than 0.001 to a maximum of 100 iterations. The log-likelihood of the parameters given the data is computed for the parameters obtained using each of the algorithms. The RMS error of the difference between the learned parameters and the ground truth is then computed and used as a metric to evaluate the results of EM and \emph{Discrete}. 
\end{subsection}

\begin{subsection}{Simple Bayesian Network Results}
The procedure described above was applied 30 times to the first kind of Bayesian network, and the results are on Fig.\ref{res1} and Fig.\ref{res2} as boxplots with median, as well as first and third quartiles. The network is exactly the same in both cases, except that the normal distributions of the sensors have standard deviation 1 and 3, respectively.

%

Focusing on the first case, it can be seen that, independently of the database size, EM very slightly outperforms the \emph{Discrete} (the RMS error is reduced by less than 2\%) algorithm and, as expected, the error is much smaller when more information is provided.
 
However, the second case has sensors with bigger variance and hence it is not as straightforward to infer the discrete values from the sensors. Thus, both algorithms perform worse than they did with the first case. However, now EM clearly outperforms \emph{Discrete} (the RMS error is reduced by 58.5\% and 84.6\% with 300 and 3000 cases); moreover, the EM algorithm greatly benefits from the increase of information whereas \emph{Discrete} has similar errors on both situations.
\end{subsection}

\begin{subsection}{Complex Bayesian Network Results}

\begin{figure}
\centering
\subfigure[]{\includegraphics[width=0.8\columnwidth]{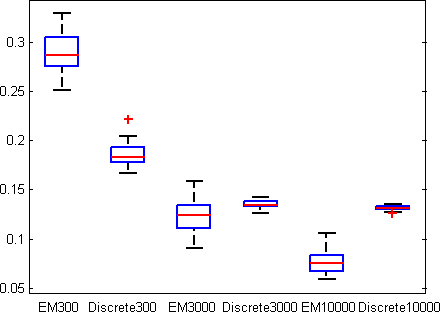}\label{res3}}
\vspace{8pt}%
\subfigure[]{\includegraphics[width=0.8\columnwidth]{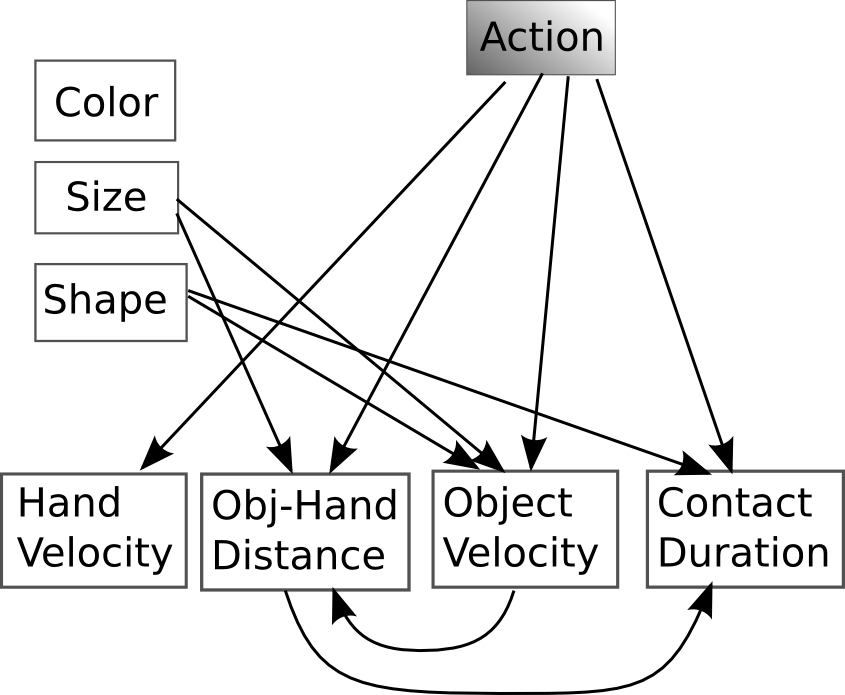}\label{graph3}}
\caption{a) RMS error of the difference between learned parameters and the real ones given a database with 300, 3000 or 10000 cases. b) Discrete variables of the realistic Bayesian network. The continuous variables (sensors) have been removed for clarification.}
\end{figure}

\begin{figure}
\centering
\includegraphics[scale=0.5]{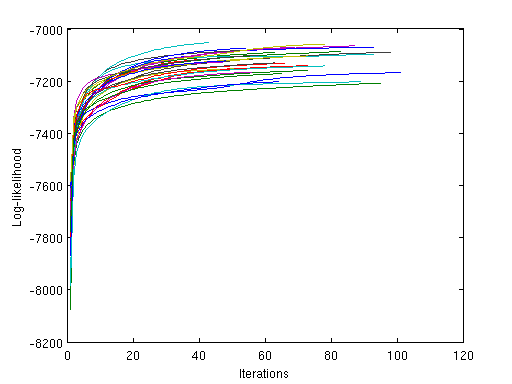}
\caption{Evolution of the log-likelihood of the parameters for the realistic network with a database of 300 cases}\label{logs}
\end{figure}

Apart from these tests with the simple Bayesian network, similar tests were done with the more complex (125 degrees of freedom) network. The means of the sensors distributions are the same as before, the standard deviations are 3, as in the second case. This time around, the procedure was applied exactly as described 30 times, with database sizes of 300 and 3000, and additionally, with a database size of 10000. The results of this procedure are summarized on Fig.\ref{res3}.

Just looking at the results for the database size of 300, one could claim that, when applied to this more complex network, the \emph{Discrete} algorithm is superior to EM. However, given that the network has 125 degrees of freedom, it is very hard to believe that any meaningful information can be extracted from only 300 experiences, that would allow the inference of the parameters of the model. As can be seen in Fig.\ref{logs}, with such a short amount of experiments, even though the log-likelihood of the parameters converges, the result is still far from good.

This suspicion is confirmed, and the trend that was observed on the previous case is seen again when the 3000 and 10000 database size results are taken into account. On the 3000 case, the RMS error for the \emph{Discrete} is slightly lower but for the EM algorithm it becomes less than half (compared to the 300 case), the RMS error is 7.7\% lower for the EM than for the \emph{Discrete}. When one observes the 10000 database size, it becomes evident that EM makes use of the extra information to perform a much better inference (the RMS error is reduced by 43.1\%), while the \emph{Discrete} algorithm performs almost at the same level as with the 3000 database size.

\end{subsection}
\end{section}

%% file: concfuture.tex
\section{Conclusions and Future Work}

It has been shown that considering the full distribution of the GMM applied to affordance learning performs better than the MAP estimate, especially in situations where there is significant overlap between the various values that descriptors of object features and effects can take. 

However, there are a number of flaws in the application of the algorithm described. Namely, its execution time of the EM algorithm is much slower than the close form solution of the network parameters using the discrete nodes directly. EM is also a batch algorithm, hence there can be no iterative learning in the sense that the algorithm update the parameters as new information becomes available. However, there are some algorithm for online-EM which could be applied in this setup \cite{Liang2009,Cappe2009}.

Interesting future work would include the development of an algorithm with comparable performance to that of EM but more efficient, particularly with regards to its time complexity dependence on the number of nodes, in order to make the algorithm scalable. This algorithm would ideally be able to store the previously performed inference as a prior, allowing iterative learning. This algorithm might also be extended to deal with the structure of the problem similarly to \cite{Friedman98}.

Also of interest is the question of whether it is feasible to acquire the amounts of information that EM appears to need in order to perform a satisfactory inference of complex Bayesian Networks, given the context, a robot that performs experiments just like a child would.

%% file: appendix.tex
\appendix
\label{app}

In this section, we derive the equations for the EM algorithm.

\subsubsection{E step} -- Compute the expected value of the complete log-likelihood with respect to the distribution of the complete data (on $K$ experiments):
\begin{align*} 
Q(\theta,\theta^{old})& = E_{p(z | y, \theta^{old})}\left[\log p(x|\theta)\right] \\
 & = \sum_{k=1}^K E_{p(z^{(k)}|y^{(k)},\theta^{old})}\left[\log p\left(x^{(k)}|\theta\right)\right] \\
 & = \sum_{k=1}^K \sum_{z^{(k)}} \log\left[p\left(y^{(k)},z^{(k)}|\theta\right)\right] \underbrace{p\left(z^{(k)}|y^{(k)},\theta^{old}\right)}_{w_k} \\
 & = \sum_{k=1}^K \sum_{z^{(k)}} w_k \left[ \log p\left(y^{(k)}|z^{(k)}\right)  + \log p\left(z^{(k)}|\theta\right) \right]   \\
 & = \underbrace{\sum_{k=1}^K \sum_{z^{(k)}} \left[w_k \log p\left(y^{(k)}|z^{(k)}\right) \right]}_\alpha +\\ 
 & + \sum_{k=1}^K \sum_{z^{(k)}} w_k \log p\left(z^{(k)}|\theta \right) \\ 
 & = \alpha + \sum_{k=1}^K \sum_{z^{(k)}} \left[ w_k \sum_{i=1}^N \log \underbrace{p\left(z_i^{(k)}|Pa_{i},\theta \right)}_{\theta_i^{(k)}} \right] 
\end{align*}
The, we can do the E step by computing
\begin{equation}
Q(\theta,\theta^{old}) = = \alpha + \sum_{k=1}^K \sum_{z^{(k)}} \left[ w_k \sum_{i=1}^N \log \theta_i^{(k)} \right]
\end{equation}

\subsubsection{M step} -- perform the maximization 
\[\theta^* = \arg \max_\theta \ Q(\theta,\theta^{old})\]
subject to the constraints $\sum_{z_i^{(k)}=1}^{r_i} \theta_i = 1$, being $r_i$ is the number of different values the $i^{th}$ node can take. 

We can use  Lagrange Multipliers $\lambda_i$ to solve the constrained optimization problem. For simplicity in the derivation, we replace $z_i^{(k)}$ with $z$.
\begin{equation}
\Lambda = Q(\theta,\theta^{old}) - \sum_{i=1}^n \sum_{Pa_{i}} \lambda_i \left(\sum_{z=1}^{r_i} \theta_i \ - 1\right)
\label{lagmulti}
\end{equation}
Now, we can derivate $\Lambda$ with respect to $\theta_i$ and make it equal to $0$. Then, developing the equation for a fixed $z_i^{(k)}$ and $Pa(z_i^{(k)}) = Pa_i$:

\begin{equation*}
-\lambda_i + \sum_{k=1}^K \sum_{\overline{Pa}_i} \frac{w_k}{\theta_i} = 0 \Leftrightarrow
\end{equation*}

\begin{equation}
\Leftrightarrow \lambda_{i} \theta_i = \sum_{k=1}^{K} \sum_{\overline{Pa}_i} w_k \Leftrightarrow \label{thetaijyi}
\end{equation}

\begin{equation}
\Leftrightarrow \lambda_{i} \sum_{z=1}^{r_i} \theta_i = \lambda_i =  \sum_{z=1}^{r_i} \sum_{k=1}^{K} \sum_{\overline{Pa}_i} w_k \label{lambdaij}
\end{equation}

Replacing \eqref{lambdaij} in \eqref{thetaijyi}:

\begin{equation} 
\theta_i =\frac{\sum_{k=1}^{K} \sum_{\overline{Pa}_i} \ w_k}{\sum_{z=1}^{r_i} \sum_{k=1}^{K} \sum_{\overline{Pa}_i} \ w_k} \label{eq:theta}
\end{equation}
Thus, we can directly estimate the parameters $\theta_i$. Furthermore, we have defined $w_k$ such us
\[
  w_k \equiv p\left(z^{(k)}|y^{(k)},\theta^{old}\right)
\]
where, we can integrate over all hiden variables that are not parents of $z_i^{(k)}$. Thus,
\begin{equation}
 \sum_{\overline{Pa}_i} \ w_k = p\left(z_i^{(k)},Pa_i|y^{(k)},\theta^{old}\right)
\label{sumtheta}
\end{equation}
By replacing \eqref{sumtheta} on \eqref{eq:theta} we obtain equation \eqref{theta}.